# Data-Driven Investigative Journalism For Connectas Dataset

**Aniket Jain**
H John Heinz III College
Carnegie Mellon University
Pittsburgh, PA 15213
*avjain@andrew.cmu.edu*

**Bhavya Sharma**
H John Heinz III College
Carnegie Mellon University
Pittsburgh, PA 15213
*bsharma@andrew.cmu.edu*

**Paridhi Choudhary**
H John Heinz III College
Carnegie Mellon University
Pittsburgh, PA 15213
*paridhic@andrew.cmu.edu*

**Rohan Sangave**
H John Heinz III College
Carnegie Mellon University
Pittsburgh, PA 15213
*rsangave@andrew.cmu.edu*

**William Yang**
School of Computer Science
Carnegie Mellon University
Pittsburgh, PA 15213
*wyang1@andrew.cmu.edu*

## Abstract

The following paper explores the possibility of using Machine Learning algorithms to detect the cases of corruption and malpractice by governments. The dataset used by the authors contains information about several government contracts in Colombia from year 2007 to 2012. The authors begin with exploring and cleaning the data, followed by which they perform feature engineering before finally implementing Machine Learning models to detect anomalies in the given dataset.

## 1   Introduction

Investigative journalism has been defined by David Spark in his book [1] as a means for "examining how the opportunities for investigations can be found and pursued, how informants can be persuaded to yield needed information and how and where this information can be checked". [1] Investigative journalism is applied in many fields including crime, corruption, frauds, and government official malpractice [1]. This paper dives into the cases of corruption in Colombia and uses data provided by Connectas, a non-profit journalism project, to discover ways of using machine learning algorithms for detecting fraudulent cases.

Colombia has been a constant target of rampant corruption cases. Despite multiple efforts to tackle the issue Colombia still ranks poorly (96 out of 180 countries) in terms of 'perceived levels of corruption in the public sector' [2]. Various cases of malpractice in government came to light in the recent past such as the case of construction giant Oderbrecht which admitted to paying $788 million in bribes to win contracts from 2001 to 2016. Another high profile example is of the former senator, Ivan Moreno, who was sentenced to prison in 2014 for abetting the embezzlement of approximately $500 million from public procurement contracts. [3]

In the light of these events, several investigative journalists have begun a crusade to uncover such cases and bring the culprits to justice. Connectas is one such organization which uses investigative journalism to reveal hidden truths behind corrupt government contracts. [4] One of their tasks is working towards the "dissemination of information on key issues for the development of the Americas." [4]

To learn of effective ways of investigating cases, Connectas reached out to Auton Lab at Carnegie Mellon University and provided us with a dataset of Colombian government contracts. The primary task of the authors is to help Connectas understand the data better and provide them with a methodology to discover anomalies, streamline and improve the process of identifying the cases of corruption.

The dataset includes details of all contracts made by the government to purchase goods, services, and works between 2007 and 2012 in Spanish. We aim to leverage natural language processing techniques and different one-class classification algorithms for anomaly detection in data.

## 2     Data cleaning and exploratory analysis

The very first step of our analysis was to explore and understand what the data contains. The data provided by Connectas was completely in Spanish making it difficult to comprehend. Therefore, we started with translating the column labels to English in order to understand the coverage of data. The attributes in the data and their interpretation according to us are mentioned below:

Table 1: Translation of labels

| Attribute Name | Interpretation |
| --- | --- |
| NIVEL | Level of the project: Territorial/National? |
| ORDEN | Order: Further level of the contract |
| NIT_ENTIDAD | NIT of Entity |
| NOMBRE_ENTIDAD | Entity Name |
| TIPO_MODALIDAD | Type of Modality |
| NUMERO_CONSTANCIA | Constance Number |
| ID_OBJETO_CONTRATO | Contract ID |
| OBJETO_CONTRATO | Contract Name |
| DETALLE_OBJETO | Detailed description of a contract |
| TIPO_CONTRATO | Contract Type |
| CUANTIA | Amount of Loan Sanctioned |
| VALOR_DEFINITIVO | Definitive Value |
| FECHACREACION | Date on which the contract was created |
| FECHAESTADOBORRADOR | Date on which state of contract was drafted |
| FECHAESTADODESCARTADO | Date on which the state of contract was |

|  | described |
|---|---|
| FECHAESTADOCONVOCADO | Date on which the state of contract was checked |
| FECHAESTADOADJUDICADO | Date on which state of contract was awarded |
| FECHAESTADOTERMANORMALDESPCONV | STATE.DATE.ABNORMAL.TERM.OF.SPCONV |
| FECHAESTADOTERMANORMALDESPCONV_1 | STATE.DATE.ABNORMAL.TERM.OF.SPCONV.1 |
| FECHAESTADOTERMANORMALDESPCONV_2 | STATE.DATE.ABNORMAL.TERM.OF.SPCONV.2 |
| FECHAESTADOTERMANORMALDESPCONV_3 | STATE.DATE.ABNORMAL.TERM.OF.SPCONV.3 |
| ESTADO_PROCESO | State of the process |
| NOMBRE_CONTRATISTA | Contractor's Name |
| NIT_CONTRATISTA | Contractor's NIT |
| FECHA_FIRMA_CONTRATO | Date on which the contract was signed |
| VALOR_CONTRATO | Contract Value |
| VALOR_ADICIONES | Additional Value Sanctioned |
| VALOR TOTAL | Total Value of the contract |

### 2.1 Interpretation

The interpretation of attributes related to the contracts is based on literal translation and expert opinions, but we are not aware of the relevance and the meaning of different columns containing information about contract amount. Hence, we are assuming that CUANTIA (amount of loan sanctioned) and VALOR_DEFINITIVO (definite value) are relevant for our analysis and we are interpreting them as the 'total amount sanctioned' and 'definitive value' of total amount withdrawn.

We expected these columns to reflect that the 'definitive value' is less than or equal to 'total amount sanctioned.' But we observed that at a certain point the 'definitive value' is greater than the 'total amount sanctioned.' The difference between the columns can hence be an indicator of whether excess money was withdrawn against a contract. Thus, we added an additional feature to reflect the difference between the two columns.

They also had a 95% correlation which we believe could be because of data error or our interpretation of the columns being wrong. (See Figure 1, Page 4)

The other peculiar thing about the dataset was that each contract with the same ID_OBJETO_CONTRATO (Contract ID) and FECHACREACION (creation date) was entered multiple times with different amounts. We removed such duplicate records and kept only the first entry as the representative. There were some entries with the 'creation dates' later than FECHAESTADOADJUDICADO (the date of signing the contract) which seems suspicious. The reason behind this is unclear and unknown to experts as well. We created a feature to indicate whether this behavior was observed in any contract.

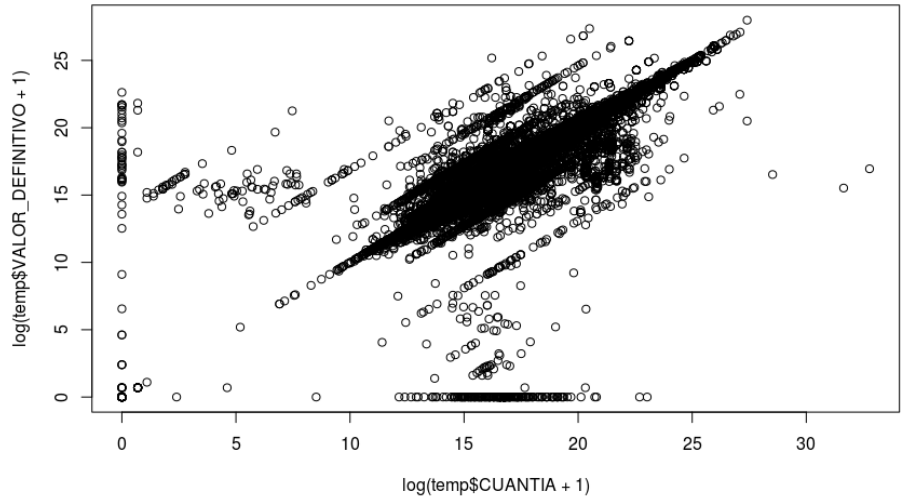

Figure 1: Correlation between 'Valor_Definitivo' and 'Cuantia'

The amounts also showed a very high spread because of which the distribution of amounts was not clear. We transformed all the 'VALOR' (value) columns using log transformation to convert them into normally distributed variables.

## 3   Feature creation using Natural Language Processing

The dataset contained two columns which described the contracts in Spanish test – OBJECTO_CONTRACTO (contract object describing the service provided) and DETALLE_OBJETO (detailed description of the contract). In order to use a machine learning models on the dataset, it was important to create features from the text columns.

We used a 'Bag of Words' approach to identifying relevant words as features and recorded their frequencies. We used Scikit-learn's CountVectorizer [5] to tokenize a corpus of text and count frequency of each word in each document. To identify features efficiently, we removed 'stopwords' (frequently occurring Spanish words that do not provide information) from our analysis. Stopwords in the Spanish language were identified using NLTK module.[6]

We identified roughly 400 features from the summary text and limited the maximum features allowed by the algorithm to be 10000 to give us only the relevant features. We also limited the vectorizer only to pick up words that have appeared in more than 0.01 % of the documents and those that have appeared in less than 50% of the documents to remove very common and very rare words from the analysis. All these steps were precautionary measures to remove any bias in the text features.

## 4   Modeling

Once the features were created, the dataset was ready for modeling. The primary task was to identify the anomalies in the dataset to give Connectas a direction for investigation. We used two methods for modeling anomaly detection :

1. Regression over amount

2. Isolation Forest

## 4.1 Regression over the amount

Hypothesis: There is a pattern in the legitimate contracts between CUANTIA (amount of loan sanctioned) and VALOR_DEFINITIVO (definite value). Hence the amounts of corrupt deals will not comply with these patterns.

Upon performing correlation analysis for amount sanctioned ('CUANTIA') and amount utilized ('VALOR DEFINITIVO'), we had found that there is a strong pattern for most of the records. However, there are sufficient data points that could be outliers. To identify if a contract had utilized unusually more amount, we implemented regression over the amount. By predicting the amount for a contract, we could find the difference between what the usual utilization should be and what the contract utilized. For a significantly large difference, this could be an alarming signal indicating a possible corrupt contract.

After prediction, we could not explore the top contracts with highest above-said differences due to the time constraint of completing this project. We do believe that performing this analysis may reveal valuable information on why the utilized amount is unusually higher that how much it should be. It is also important to note that the outliers or unusually contracts should not be included while learning the regression model.

## 4.2 Isolation Forest

Hypothesis: There are patterns in the overall data, and the unusual contracts will stand out having none/very few patterns.

To find out unusual contracts we used Isolation Forest [7]. We fit the model overall data, which included original and engineered features of the bag of words. Isolation forest splits the data over nodes randomly rather than information gain (used by random forest). By generalizing the data over these splits, it checks for data points that split very early in the decision tree. An early split indicates the data point has some feature with a distinctive value than others.

We have extracted nine anomalous/unusual contracts from Isolation Forests. As Isolation Forest randomly split over features, it is difficult to interpret the resulting anomalies. One way we propose to do this is through labeling the anomalies and training a decision tree/random forest classifier to classify them. This method not only gives the underlying feature importance for a contract to be corrupt potentially but also can be used for future classification of contracts.

## 5 Results from Isolation Forest

The Isolation Forest model was implemented at two stages in the process. First before the inclusion of DETALLE_OBJETO (contract description) using bag of words approach. Second including the bag of words. The types of contracts and their mode for the anomalies are given in Table 2 and Table 3 on Page 6.

We hypothesized that these implemented tools would highlight anomalous/unusual contracts that could potentially be lying in the Construction/ Food Provision services. The lack of expert knowledge and strong assumption of an outlier being a "corrupt" contract, has limited us from validating our hypothesis.

However, we believe that our tool would be useful for the investigative journalists to:
   1. Explore and represent the data well
   2. Utilize all information possible to assist outlier detection

Table 2: Contract types and modalities without Bag of Words

| Without Bag of Words | |
|---|---|
| Number of Contracts | 9 |
| TYPE OF CONTRACT | TYPE MODALITY |
| Compraventa | Seleccion Abreviada de Menor Cuantia (Ley 1150 de 2007) |
| Compraventa | Licitacion Publica |
| Suministro | Licitacion Publica |
| Suministro | Seleccion Abreviada de Menor Cuantia (Ley 1150 de 2007) |
| Obra | Licitacion Publica |
| Suministro | Subasta |
| Compraventa | Subasta |
| Prestacion de Servicios | Seleccion Abreviada de Menor Cuantia (Ley 1150 de 2007) |
| Suministro | Subasta |

Table 3: Contract types and modalities with Bag of Words

| With Bag of Words | |
|---|---|
| Number of Contracts | 7 |
| TYPE OF CONTRACT | TYPE MODALITY |
| Prestacion de Servicios | Contratacion Directa (Ley 1150 de 2007) |

## 6  Conclusion

This paper introduced the concept of investigative journalism and used the case of corruption in Colombia to demonstrate how Machine Learning models can be used to assist the journalists in investigating the cases. We performed the essential exploratory analysis, feature engineering and built models for anomaly detection. Although the analysis is not complete, it certainly builds a base and provides direction for the journalist to move ahead in the cases. Our demonstration in this paper is an example of how Investigative Journalism can benefit from Machine Learning and it explores the numerous possibilities of extracting useful insights from the data provided.

## 6  Acknowledgments

We would like to thank Auton Lab, CMU for conducting the HackAuton 2018 and giving us the opportunity to work with an interesting and valuable dataset. We would also like to thank Connectas for providing and trusting AutonLab with the sensitive data about government contracts in Colombia.